\documentclass[11pt,a4paper]{article}
\usepackage{times}
\usepackage{latexsym}

\usepackage{url}
\usepackage{graphicx}
\usepackage{xcolor}
\usepackage{amsmath}
\usepackage{subcaption}

\usepackage[draft]{todonotes}
\usepackage[hyperref]{emnlp}

\aclfinalcopy 

\title{Simple, Scalable Adaptation for Neural Machine Translation}

\author{Ankur Bapna \qquad
  Naveen Arivazhagan \qquad
  Orhan Firat\qquad \\
  {\centering Google AI} \\
  {\centering \tt {\{ankurbpn,navari,orhanf\}}@google.com}
  }

\date{}

\begin{document}
\maketitle
\begin{abstract}

Fine-tuning pre-trained Neural Machine Translation (NMT) models is the dominant approach for adapting to new languages and domains. However, fine-tuning requires adapting and maintaining a separate model for each target task. We propose a simple yet efficient approach for adaptation in NMT. Our proposed approach consists of injecting tiny task specific adapter layers into a pre-trained model. These lightweight adapters, with just a small fraction of the original model size, adapt the model to multiple individual tasks simultaneously.

We evaluate our approach on two tasks: (i) Domain Adaptation and (ii) Massively Multilingual NMT. Experiments on domain adaptation demonstrate that our proposed approach is on par with full fine-tuning on various domains, dataset sizes and model capacities. On a massively multilingual dataset of 103 languages, our adaptation approach bridges the gap between individual bilingual models and one massively multilingual model for most language pairs, paving the way towards universal machine translation.

\end{abstract}

\section{Introduction}
Recent developments in deep learning have led to significantly improved quality on Neural Machine Translation (NMT) \citep{kalchbrenner-blunsom:2013:EMNLP,sutskever2014sequence,bahdanau2014neural,DBLP:journals/corr/VaswaniSPUJGKP17}. While NMT performance on sentence level translation for high resource languages seems to be dramatically improved \citep{wu2016google,hassan2018achieving}, performance on out-of-domain data or low resource languages, still remains pretty poor \citep{duh2013adaptation,koehn2017six,farajian2017multi,dong2015multi,zoph2016transfer}. This has generated significant interest in adaptation approaches, that can leverage the huge amounts of parallel data available for high resource languages, to improve translation performance on low resource tasks. In this work we focus on two adaptation tasks: (i) Domain Adaptation, to improve the performance of in-domain translation by leveraging out-of-domain datasets \citep{Luong-Manning:iwslt15,freitag2016fast}, and (ii) Multilingual NMT, to improve the translation quality on low resource languages via co-training with similar languages \citep{dong2015multi,firat2016multi,johnson2016google,zoph2016transfer,neubig2018rapid,DBLP:journals/corr/abs-1903-00089,arivazhagan2019massively}.

\begin{figure*}[h!]
\begin{center}
\includegraphics[scale=0.5]{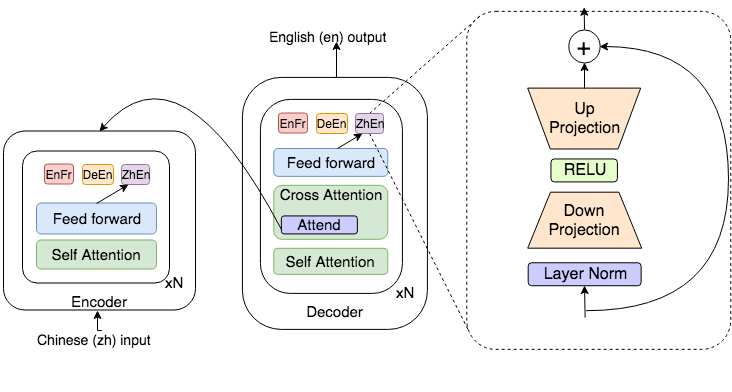}
\caption{Diagrams depicting (i) Left: the proposed layout of a Transformer enhanced with language specific adapters (ii) Right: the architecture of each residual adapter layer. While the figure depicts use of adapters for multilingual NMT, the same formulation can be used for domain adaptation.}
\label{fig:resadap}
\end{center}
\end{figure*}

While several approaches have been explored in literature \citep{chu2018survey}, full fine-tuning of model parameters remains the dominant approach for adapting to new domains and languages \citep{Luong-Manning:iwslt15,neubig2018rapid}. However, fine-tuning requires training and maintaining a separate model for every language, for every domain. As the number of languages and domains grow, training, maintaining and serving a separate model for every task becomes infeasible. This is further exacerbated by the increasing model capacity of state-of-the-art NMT models \citep{shazeer2018meshtensorflow,bapna2018training,huang2018gpipe}; full fine-tuning is just too parameter inefficient. In addition to the growing number of models, fine-tuning requires very careful hyper-parameter tuning (eg. learning rate, regularization knobs etc.) during adaptation, and is prone to rapid over-fitting \citep{Sennrich_2016,D17-1156}. This sensitivity to hyper-parameters and over-fitting to the adaptation corpus become worse in the setting of high capacity models.

These weaknesses beckon the need for parameter-efficient, scalable and hyper-parameter insensitive approaches for adaptation. The ideal adaptation approach should also offer the flexibility to adapt to tasks of varying complexity and adaptation corpus sizes, within a single model.

In this work we propose using light-weight adapter layers, which are transplanted between the layers of a pre-trained network and fine-tuned on the adaptation corpus. 
Adapting only the light-weight layers enables our approach to be parameter efficient, and eases the scalability of the approach to large models. The capacity of these adapters can be adjusted to match the requirements of the target task, making them suitable for a variety of adaptation tasks. By separating the parameters of the original network and each adaptation task, our approach circumvents catastrophic interference \citep{mccloskey1989catastrophic} with the original model parameters, and allows us to simultaneously adapt a single model to multiple domains and languages, while retaining the quality on the source languages and domains.


We make three major contributions in this work: (i) we propose a formulation of adapter layers for NMT adaptation that enables us to tune their capacity according to the target task complexity and corpus size,
(ii) we evaluate our approach on domain adaptation, and demonstrate that light-weight adapters match the performance of full fine-tuning based adaptation at a fraction of the per-domain parameter cost, and
(iii) we use adapters to train a massively multilingual model on 103 languages, and demonstrate that it is possible to train a single model that significantly improves transfer performance on low resource languages, without huge regression on high resource language pairs.
 
By demonstrating the effectiveness of adapters on domain adaptation and massively multilingual translation, we make progress towards a flexible universal translation model for all languages and domains.

\section{Related Work}
Several approaches have been proposed in recent literature that try to address the shortcomings of full fine-tuning when applied to domain adaptation \citep{chu2018survey}.
\citet{michel2018extreme} proposed a space efficient approach to adaptation that introduces domain-specific biases to the output vocabulary, enabling extreme personalization in settings where small amounts of data are available for a lot of different domains.
\citet{thompson2018freezing} fine-tune selected components of the base model architecture, in order to determine how much fine-tuning each component contributes to the final adaptation performance.
\citet{wuebker2018compact} propose introducing sparse offsets from the base model parameters for every domain, reducing the memory complexity of loading and unloading domain specific parameters in real world settings. \citet{bapna2019non} train the base model to utilize neighboring samples from the training set, enabling the model to adapt to new domains without the need for additional parameter updates.
Learning Hidden Unit Contribution (LHUC) \citep{vilar2018learning} is perhaps closest to our work in spirit. They introduce domain specific gates that control the contribution of hidden units feeding into the next layer. However, they introduce a limited amount of per-domain capacity which doesn't scale well when a lot of domain specific data is available.

Residual Adapters were first introduced for adapting vision models in \citet{rebuffi2017learning}, but their formulation used a single projection layer, without any tunable hyper-parameters that could be used to adjust capacity based on the target domain. \citet{houlsby2019parameterefficient} utilized a new formulation of adapters to adapt BERT \citep{devlin2018bert} to multiple tasks simultaneously. Our formulation of adapters is motivated by theirs, but differs in a few respects. \citet{houlsby2019parameterefficient} introduce adapters after every sub-layer (self-attention, feed-forward) within a transformer layer, and re-train existing layer normalization parameters for every new domain. We simplify this formulation by leaving the parameters frozen, and introducing new layer normalization parameters for every task, essentially mimic-ing the structure of the transformer feed-forward layer.

\section{Approach}

Our approach consists of two phases: (i) Training a generic base model, and (ii) adapting it to new tasks with added small network modules. We first take a standard NMT model which is trained on a large source corpus. Following convergence,\footnote{We leave the methodology for defining convergence task specific, e.g. early stopping on validation set accuracy or number of training steps.} all model parameters are frozen, preserving the information learned during this pre-training phase.
Next, per-task light-weight adapter layers (see Fig.~\ref{fig:resadap} right pane) are introduced after every layer in the encoder and the decoder (see Fig.~\ref{fig:resadap} left pane). We fine-tune the parameters of these task-specific adapters on the adaptation corpus. This procedure can be followed for every additional task, allowing us to train a single model for all tasks simultaneously, with a small set of task-specific adapters.

\paragraph{Adapter Modules} Our design principles for adapter modules are simplicity and flexibility. We propose a simple single hidden-layer feed-forward network formulation for adapters, with a non-linear activation function between the two projection layers. The inner dimension of these two projections is the only knob to tune. This allows us to adjust the capacity of the adapter module easily, depending on the complexity of the target task or domain. Additionally, we normalize the input of the adapters, in order to make the module plug-able into any part of the base network, irrespective of the variations in the activation patterns/distributions. This parametrized normalization layer allows the module to learn the activation pattern of the layer it's injected into. Finally, to allow the adapter module to represent a no-op if necessary, we wrap it with a residual connection.

\paragraph{Formulation} Let $z_i$ be the output of the $i$-th layer, of dimension $d$. We first apply layer-normalization \citep{ba2016layer} to the inputs of the adapter corresponding to task $T$.
\begin{equation}
    \tilde{z}^{T}_i = LN_T(z_i).
\end{equation}

\noindent This is followed by a projection layer of dimension $b$. The dimension of the projection can be tuned based on the task complexity and the size of the adaptation corpus. This now allows us to make it a bottleneck layer for compression, or over-parametrize it with a dimension larger than the input dimension.\footnote{Following the wiring choices of Transformer feed-forward network \citep{DBLP:journals/corr/VaswaniSPUJGKP17}.}

\begin{equation}
    h^T_i = relu (W^{T}_{bd}\tilde{z}^{T}_i).
\end{equation}

\noindent Lastly, the inner representation is projection back to the input dimension $d$, and combined with a residual connection \citep{he2015deep}:

\begin{equation}
     x_i^{T} = W^T_{db}h^T_i + z_i.
\end{equation}

 Our proposed formulation for adapters, and their incorporation into Transformers \citep{DBLP:journals/corr/VaswaniSPUJGKP17} is illustrated in Figure~\ref{fig:resadap}. This self-contained adapter module can be injected between any two layers of the network, without disrupting the original operation.

\section{Domain Adaptation}
We first compare the adaptation performance of the light-weight residual adapters against full fine-tuning and LHUC \citep{vilar2018learning} on a large scale English-French domain adaptation task. 
\subsection{Dataset}

We use the WMT En-Fr training set (36M pairs) as our out-of-domain (source) training corpus. NMT models trained on WMT are then adapted to the IWSLT'15 En-Fr corpus, consisting of 237k sentence pairs. We also evaluate adaptation performance on the JRC-Acquis dataset \footnote{http://opus.nlpl.eu/JRC-Acquis.php}, which is an extremely narrow domain dataset containing 797k sentence pairs in the training split. For IWSLT, we use the test corpora from 2012-14 for validation, and the test corpus from 2015 as the test set. For JRC-Acquis the test and validation set contain 6574 and 5121 sentence pairs respectively. We also evaluate the translation performance of the non-adapted base model on newstest-2014.

\begin{table}[ht]
\begin{center}
  \begin{tabular}{l|l|l|l|l}
    \hline
    Dataset & Base & FT & LHUC & Adap.\\
    \hline \hline
    WMT'14 & 42.80 & - & - & -\\ \hline
    IWSLT'15 & 41.33 & \textbf{44.59} & 43.33 & \textbf{44.63}\\\hline
    JRC & 54.60 & \textbf{64.13} & 57.10 & 63.48\\
    \hline
  \end{tabular}
\caption{Domain adaptation performance with different adaptation strategies. Base refers to the baseline NMT model trained on the WMT'14 En-Fr training corpus. FT refers to the fine-tuning upper bound, adapting all the model parameters by incrementally training on in-domain training data. LHUC adds additional task-specific gating parameters to the pre-trained model, which are trained on the in-domain data, as described in \citet{vilar2018learning}. Adap. is the proposed adaptation approach, adding domain specific adapter layers trained on the in-domain training corpus. \label{tab:adap}}
\end{center}
\end{table}

\subsection{Using Adapters for Domain Adaptation}
\label{subsec:adap_domadap}
When using adapters for domain adaptation, we follow the following two step approach:
\begin{itemize}
\item Pre-training: Pre-train the NMT model on a large open-domain corpus. Freeze all the parameters of this pre-trained model.
\item Adaptation: Inject a set of domain-specific adapter layers for every target domain. These adapters are then fine-tuned to maximize performance on the corresponding domains. This step can be applied any time a new domain is added to the model.
\end{itemize}
As discussed in Section~\ref{subsec:adap_multi}, we follow a slightly different approach when using adapters for multilingual NMT.

\subsection{Models and Hyper-parameters}
We use a larger version of Transformer Big containing 375M parameters as our base model. Our model is identical to \citet{DBLP:journals/corr/VaswaniSPUJGKP17}, having 6 encoder and decoder layers (12 in total), except that we use hidden layers of size 8192 instead of 4096, and a learning rate schedule of (3.0, 40K)\footnote{(3.0, 40K) schedule is the shorthand for a learning rate of 3.0, with 40K warm-up steps for the schedule, which is decayed with the inverse square root of the number of training steps after warm-up.}, following \citet{chen2018best}.

\begin{figure}[t!]
\includegraphics[scale=0.35]{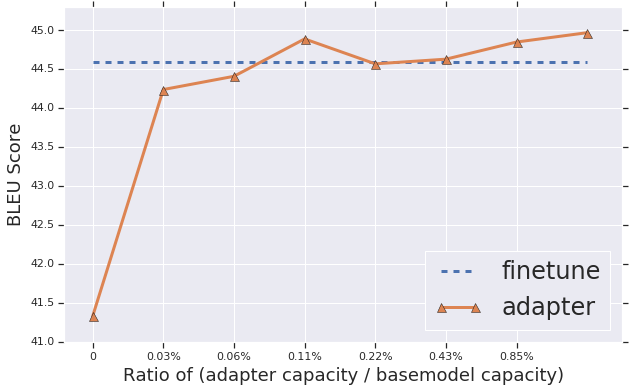}
\caption{IWSLT Adaptation performance vs adapter capacity in terms of percentage of the base model's capacity. The range of model capacity plotted here corresponds to adapter bottleneck dimensions of 4, 8...256 respectively.}
\label{fig:iwslt}
\end{figure}

For full fine-tuning, we continue training on the in-domain corpus without resetting the optimizer accumulators or the learning rate schedule. This allows us to fine-tune gradually, avoiding rapidly over-fitting to the target domain. We also conducted fine-tuning experiments with SGD, but report results with the approach described above, based on better performance on the validation sets.
When adapting with LHUC or light-weight adapters, we train using the same learning rate schedule and optimizer used during pre-training, but restart from the 0-th step, resetting the optimizer accumulators. BLEU scores are computed on the checkpoint with the best validation performance, on tokenized, true-cased output and references using \textit{multi-bleu.perl} from Moses.
All our experiments were performed using the open source Tensorflow Lingvo \citep{lingvo} framework.

\subsection{Results and Analysis}
The results of our adaptation experiments are documented in Table~\ref{tab:adap}. On both, IWSLT and JRC-Acquis, full model fine-tuning (Full-FT columns in Table~\ref{tab:adap}) on in-domain data significantly improves translation performance compared to the base, non-adapted Transformer Big by a huge margin, 3 BLEU points for IWSLT and 9 BLEU points for JRC. LHUC also improves performance over the base model, but lags behind a fully fine-tuned model for both domains and model capacities.

On IWSLT, adapters match the performance of the fine-tuning upper bound within error margins, while adding less than 0.11\% of the original model parameters. On JRC-Acquis adapters recover around 90\% of fine-tuning improvements without updating any existing parameters, while adding around 13.5\% additional parameters.

\begin{figure}[t!]
\includegraphics[scale=0.35]{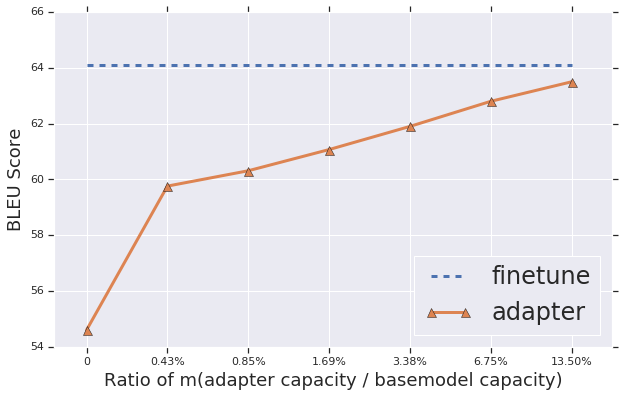}
\caption{JRC-Acquis Adaptation performance vs adapter capacity in terms of percentage of the base model's capacity. The range of model capacity plotted here corresponds to adapter bottleneck dimensions of 64, 128...2048 respectively.}
\label{fig:jrc}
\end{figure}

To demonstrate the flexibility of our approach, we quantify the trade-off between adapter capacity and adaptation performance on both IWSLT and JRC-Acquis. In Figures~\ref{fig:iwslt} and~\ref{fig:jrc}, we plot the adaptation performance on IWSLT and JRC-Acquis respectively, while varying adapter capacity. On IWSLT, we notice that residual adapters reach within 0.5 BLEU of the full fine-tuning upper bound with just 0.03\% of the model capacity, corresponding to a hidden dimension of size 4. By increasing capacity further we were able to improve over the full fine-tuning baseline by around 0.5 BLEU. On the other hand, on JRC-Acquis, adapter capacity had to be increased up to 13.5\% of the total model capacity, corresponding to a hidden dimension of size 2048, before we were within 0.5 BLEU of the full fine-tuning performance. This highlights a key strength of the approach: by varying adapter capacity it is possible to adapt the same model to domains of varying complexity and amounts of data.

\begin{figure}[h]
\includegraphics[scale=0.35]{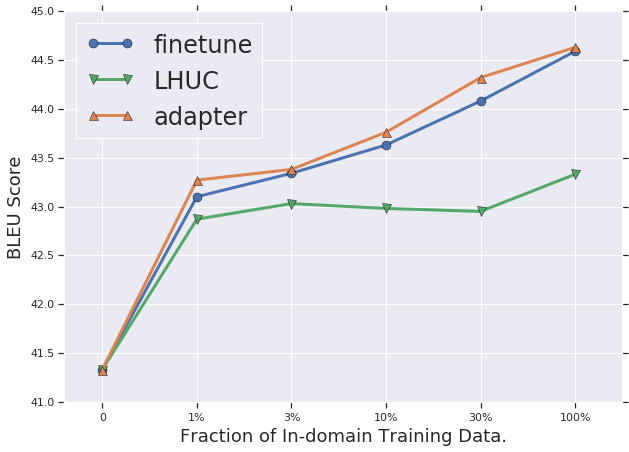}
\caption{IWSLT Adaptation performance vs fraction of the in-domain training corpus used for adaptation. The blue solid line plots the performance of fine-tuning. The red dotted line tracks the performance of LHUC, while the yellow dashed line tracks the performance of the proposed residual adapter based approach with varying amounts of training data.}
\label{fig:data}
\end{figure}

To evaluate the effectiveness of adapters when adapting with small in-domain corpora, we further compare the performance of adapters with fine-tuning on varying amounts of training data. In Figure~\ref{fig:data} we plot the adaptation performance on IWSLT, when using different fractions of the training corpus for adaptation. While LHUC is competitive with full fine-tuning and light-weight adapters for extremely small fractions, the lack of capacity limits the applicability of the approach when larger quantities of adaptation data are available. On the other hand, by tuning the capacity of adapters to match the requirements for the adaptation corpus size, we are able to match and out-perform fine-tuning on almost all evaluated data-points.

\begin{figure}[h]
\includegraphics[scale=0.35]{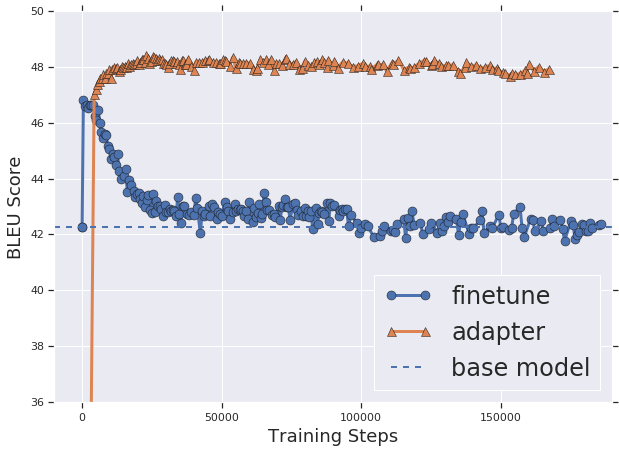}
\caption{IWSLT dev performance vs number of in-domain adaptation steps when adapting with fine-tuning vs adapters.}
\label{fig:ftva}
\end{figure}

In order to monitor the learning behavior of light-weight adapters, we compare the validation BLEU scores during the course of the fine-tuning process. Figure~\ref{fig:ftva} illustrates the comparison of the two approaches, full fine-tuning and light-weight adapters. We notice that for a reasonably small adapter size, adapter performance gradually converges to its peak and stays steady, with almost no over-fitting for a long enough period, easing final model selection. On the other hand, with full fine-tuning, optimal model selection becomes challenging due to rapid over-fitting on the adaptation corpus. This, in fact, can be remedied by carefully tuning the learning rate (and/or batch size) during adaptation, but is not trivial and needs to be done individually for every different domain, model and corpus size, favoring the simplicity of our proposed approach. 

\section{Massively Multilingual Machine Translation}
To stress test our adapters based approach, we apply this to a massively multilingual translation task on a real world dataset \citep{arivazhagan2019massively}. Most previous literature in multilingual NMT focuses on improving the performance of low resource languages \citep{zoph2016transfer,firat2016multi,neubig2018rapid}, often ignoring the source language performance of the adapted model. However, the goal of our work is to enable training a single model for all language pairs, in order to get benefits of transfer on low resource language pairs, without losing performance in the high resource setting.

\subsection{Dataset}
To highlight the flexibility of an adapters based approach, we study multilingual NMT on a massive scale, using a corpus generated by crawling and extracting parallel sentences from the web. Our corpus contains parallel documents for 102 languages, to and from English, containing a total of 25 billion sentence pairs \citep{arivazhagan2019massively}.\footnote{Limited to approximately this amount for experimentation.} The number of parallel sentences per language in our corpus ranges from around 10s of thousands to almost 2 billion. Figure \ref{fig:data} illustrates the data distribution across languages for all 102 languages studied in this paper.

\begin{figure*}[t!]
\begin{center}
\includegraphics[scale=0.4]{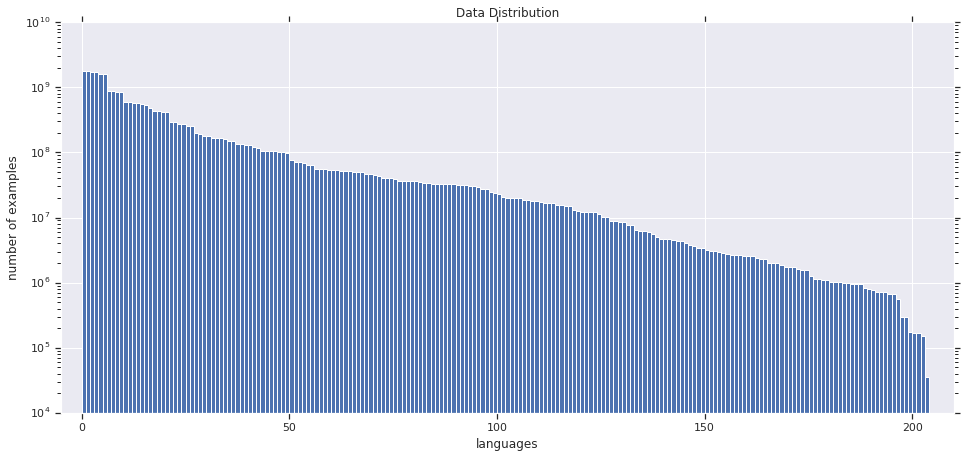}
\caption{Per language pair data distribution of the dataset used for our multilingual experiments (for 102 languages, 204 language pairs to and from English). The y-axis depicts the number of training examples available per language pair on a logarithmic scale. Dataset sizes range from the order of $10^4$ for the lowest resource language pairs to the order of $10^9$ for the largest.}
\label{fig:data}
\end{center}
\end{figure*}

\subsection{Using Adapters for multilingual NMT}
\label{subsec:adap_multi}
Our approach to using adapters for multilingual NMT diverges from domain adaptation, owing to the differences in the two tasks. 

In the domain adaptation setting, while the input and output distributions of the adaptation domain might differ from that of the base, the set of inputs and outputs is pretty much the same. In mathematical terms, both the base and adaptation domain distributions, $D_S$ and $D_T$ respectively, are defined on the same support set $\{X, Y\}$.

On the other hand, in multilingual NMT, the support sets of different language pairs have very little overlap. In this setting, adapting a model to a new language pair without learning the embeddings and softmax parameters (which correspond to the input and output support sets) would be an extremely difficult task. Following the approach used for domain adaptation in Section~\ref{subsec:adap_domadap} might not be possible here. We modify our approach to expand the input-output distribution of our initial pre-trained model to all language pairs we are interested in supporting, i.e. we can't add any new language pairs to the model during adaptation, but we use the adaptation stage to improve performance on languages learnt during pre-training.

For multilingual NMT, we follow the following two step approach:
\begin{itemize}
    \item Global training: Train a fully shared model on all language pairs, with the goal of maximizing transfer to low resource languages. 
    \item Refinement: Fine-tuning language pair specific adapters for all high resource languages, to recover lost performance during step 1. This step can only be applied for language pairs learned during global training.
\end{itemize}

\subsection{Models and Hyper-parameters}
We first train dedicated bilingual models on all language pairs to ground our multilingual analyses. We perform all our experiments with variants of the Transformer architecture \citep{DBLP:journals/corr/VaswaniSPUJGKP17}. For most bilingual experiments, we use a larger version of Transformer Big containing 375M parameters \citep{chen2018best}, and a shared source-target sentence-piece model (SPM) \citep{kudo2018sentencepiece} vocabulary with 32k tokens. We tune different values of dropout \citep{srivastava2014dropout}, depending on the dataset size for each language pair. For most medium and low resource languages we also experiment with Transformer Base. All our models are trained with Adafactor \citep{shazeer2018adafactor} with momentum factorization, a learning rate schedule of (3.0, 40K), and a per-parameter norm clipping threshold of 1.0. For Transformer Base models, we use a learning rate schedule of (2.0, 8K). BLEU scores are computed on the checkpoint with the best validation performance, on true-cased output and references.\footnote{We used an in-house implementation of mteval-v13a.pl from Moses to evaluate BLEU scores for our multilingual experiments.}

\begin{figure*}[h]
\centering
\begin{subfigure}[b]{0.9\textwidth}
   \includegraphics[width=0.9\textwidth]{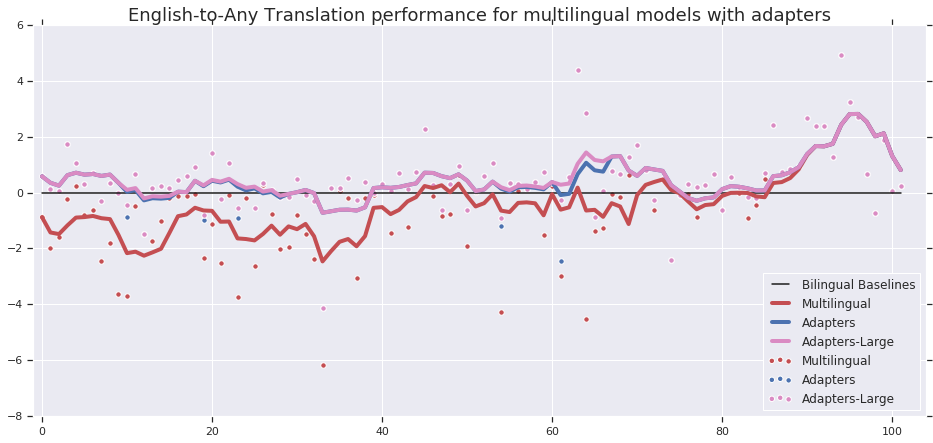}
\end{subfigure}
\begin{subfigure}[b]{0.9\textwidth}
   \includegraphics[width=0.9\textwidth]{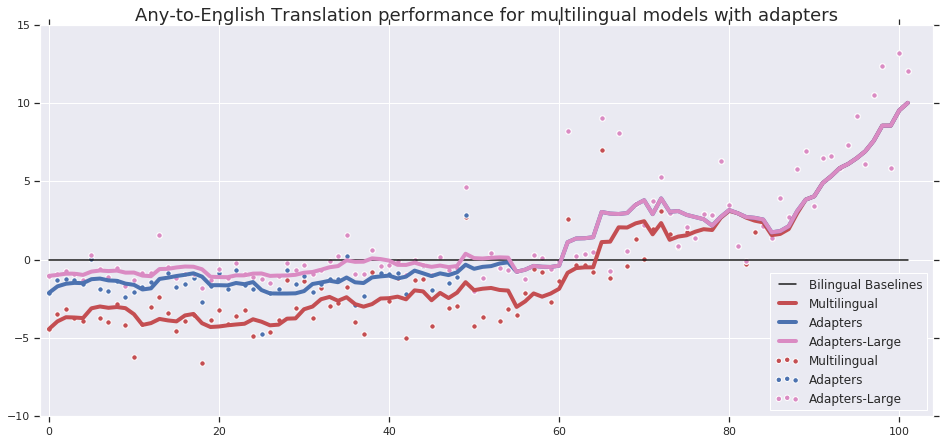}
\end{subfigure}
\caption{Trendlines depicting translation performance improvement in multilingual models with residual adapters. From left to right, languages are arranged in decreasing order of available training data. y-axis depicts the BLEU score relative to the bilingual baseline trained on the corresponding language pair. The plots correspond to the following models: (1.) Red: Multilingual model trained with sampling temperature, $T=5$ (2.) Blue: Multilingual model + Small adapters, $b=2048$ (3.) Pink: Multilingual model + Large adapters, $b=4096$. \\Note: For adapter experiments, we choose the best performance between $b=0$, $b=2048$ and $b=4096$.}
\label{fig:multi}
\end{figure*}

We now describe our approach for training the multilingual models. Due to the large imbalance in our training dataset (Figure~\ref{fig:data}), we first design a sampling strategy to simultaneously train a single model on all 204 language pairs. Sampling directly from the data distribution results in good performance on high resource languages, but low resource languages get starved. Sampling equally from all language pairs results in huge boost in low resource translation performance, but high resource languages perform significantly worse than their bilingual baselines.

To balance between high and low resource language pairs, we use a temperature based sampling strategy \citep{arivazhagan2019massively}. For a given language pair, $l_{12}$, let $D_{l_{12}}$ be the size of the available parallel corpus. Then if we sample from the union of the datasets, the probability of the sample being from language pair $l_{12}$ is $p_{l_{12}}=\frac{D_{l_{12}}}{\Sigma_{l_{12}}D_{l_{12}}}$. We set the probability of our sampled distribution to be proportional to $p_{l_{12}}^{\frac{1}{T}}$, where $T$ is the sampling temperature. Now, $T=1$ corresponds to true data distribution and $T=100$ corresponds to an (almost) equal number of samples for each language. We use $T=5$ for our multilingual model.

We train a single Transformer Big simultaneously on all 204 language pairs (102 languages to and from English), with the same hyper-parameter settings as the bilingual model. However, we use a shared SPM vocab with 64K tokens, generated using the same sampling distribution ($T=5$) used during training. We additionally use character coverage of $0.999995$ to ensure our vocab contains most of the alphabets for all 103 languages. Please refer \citep{arivazhagan2019massively} for additional training details for the base multilingual model.

Following global pre-training on all language pairs, we inject and fine-tune language pair specific adapters. The fine-tuning stage is performed separately for each language pair to reduce the device memory needed for the training process. The fine-tuned adapters can then be combined together into a single model after this stage. For fine-tuning, we use the same hyper-parameter used during global pre-training, but reset our optimizer accumulators and restart from step $0$. For our experiments we use the same bottle-neck dimension, $b=2048$, for all language pairs. This was meant to reduce the number of experiments given the large number of language pairs in our setup. For language pairs that were worse than their bilingual models after adding adapters with $b=2048$, we re-run fine-tuning with larger adapters, with $b=4096$.
In an ideal setting, the bottle-neck could be larger for the highest resource languages and $b=0$ (no adapters) for the smallest languages.

\subsection{Results and Analysis}
We plot the translation quality on different language pairs in Figure~\ref{fig:multi}. As we can see, the multilingual model significantly out-performs the bilingual baselines in the extremely low resource setting. These gains are even more amplified when translating into English, agreeing with previous work in multilingual NMT \citep{neubig2018rapid,DBLP:journals/corr/abs-1903-00089}. However, owing to the huge training corpus, we observe significant performance deterioration in the high resource languages. We attribute this deterioration to two factors: (i) Languages compete for capacity given the limited model size, and (ii) The model converges much before it trains on significant portions of the high resource datasets.

As is apparent from Figure~\ref{fig:multi}, performance on high and medium resource languages improves by huge margins after the second stage of training (adapter based refinement). Fine-tuning with adapters allows the model to see larger portions of the training data for high resource languages, and converges faster than training a model from scratch since it only updates a very small fraction of the model parameters (for most language pairs, the second stage converges within 20-50k steps, depending on the corpus size). For high resource languages, especially when translating into English, we observe further performance improvements when increasing adapter size. This again highlights the flexibility of adapters, it is possible to adjust the adapter capacity to match the complexity and resource size of the target task.

While adapters help us bridge most of the gap between bilingual and multilingual models, we still observe a minor regression for high resource languages translating into English, compared to the bilingual baselines. Although it might be possible to reduce this gap further by increasing the adapter size beyond $b=4096$, there might be more efficient ways to approach this problem, including more expressive network architectures for adapters, joint fine-tuning of adapters and global model parameters, etc. However, we leave these studies to future work.

\section{Conclusion}

In this work, we proposed \textit{light-weight adapters}, a simple yet efficient way for adapting large scale neural machine translation models. Our proposed approach, by injecting small task-specific adapter layers between the frozen base model layers during adaptation, enables the final network to adapt to multiple target tasks simultaneously, without forgetting the original parameters.

We evaluate light-weight adapters on two different adaptation tasks, domain adaptation and multilingual NMT. Experiments support the flexibility and scalability of \textit{light-weight adapters}, (i) yielding comparable or better results when compared with the standard full fine-tuning or bilingual baselines, (ii) without the need for any hyper-parameter tuning across varying adaptation dataset sizes and model capacities.

With a large set of globally shared parameters and small interspersed task-specific layers, adapters allow us to train and adapt a single model for a huge number of languages and domains. We hope that this work would motivate further research into massively multitask and universal translation models.

\section*{Acknowledgments}
We would like to thank the Google Translate and Lingvo development teams for their foundational contributions to this project. We would also like to thank Neil Houlsby, Anjuli Kannan and Yonghui Wu for helpful discussions early on in the project, and Wolfgang Macherey, Markus Freitag, Ciprian Chelba, Zhifeng Chen and the anonymous EMNLP reviewers for their insightful comments.

\bibliography{emnlp}
\bibliographystyle{acl_natbib}

\end{document}